# Stream-based Online Active Learning in a Contextual Multi-Armed Bandit Framework

**Linqi Song**
Electrical Engineering Department
University of California, Los Angeles
Los Angeles, CA 90095
songlinqi@ucla.edu

## Abstract

We study the stream-based online active learning in a contextual multi-armed bandit framework. In this framework, the reward depends on both the arm and the context. In a stream-based active learning setting, obtaining the ground truth of the reward is costly, and the conventional contextual multi-armed bandit algorithm fails to achieve a sublinear regret due to this cost. Hence, the algorithm needs to determine whether or not to request the ground truth of the reward at current time slot. In our framework, we consider a stream-based active learning setting in which a query request for the ground truth is sent to the annotator, together with some prior information of the ground truth. Depending on the accuracy of the prior information, the query cost varies. Our algorithm mainly carries out two operations: the refinement of the context and arm spaces and the selection of actions. In our algorithm, the partitions of the context space and the arm space are maintained for a certain time slots, and then become finer as more information about the rewards accumulates. We use a strategic way to select the arms and to request the ground truth of the reward, aiming to maximize the total reward. We analytically show that the regret is sublinear and in the same order with that of the conventional contextual multi-armed bandit algorithms, where no query cost is assumed.

## 1 Introduction

The active learning problem assumes that obtaining the ground truth label of a data instance is costly, and hence the algorithm needs to judiciously choose which data instance to query [1, 2]. In a stream-based online active learning setting [1, 6], an instance arrives each time, and the algorithm takes an action and chooses whether or not to request the ground truth. Most of the existing works [3, 4, 5, 7] assume that the instance is sent to an annotator and the query cost is a constant. In many applications, such as advertising, recommender systems, and computer-aided medical diagnosis, this cost can be greatly reduced by sending additional information to the annotator (such as statistical information of the reward). We extend the stream-based online active learning model by providing prior information about the ground truth to the annotator, resulting in a varying query cost. The query cost becomes lower when more accurate prior information is provided to the annotator.

Another big difference from existing active learning literature is that in our contextual multi-armed bandit (MAB) based framework, the reward depends on both the context and the arm selected. The contextual MAB, which is an extension of conventional MAB [8, 9, 13], has been widely studied recently [10, 12, 14]. The MAB model describes a sequential decision making problem, where the trade-off between the exploration and exploitation needs to be considered. The context describes the situation or environment in which a data instance arrives. In a contextual MAB framework, a context arrives at each time, and an arm is selected based on past context arrivals and realized rewards. The expected rewards are considered as a function of the arm and context. In these existing works [10, 12, 14], it is assumed that there is no cost incurred when a ground truth of the reward is requested. In contrast, we consider the query cost in our framework. Hence, each time, the action of whether or not to request a ground truth needs also to be considered by the algorithm.

Our proposed framework can be used in a wide range of application domains, including recommender systems, computer-aided medical diagnosis, advertising, etc. For example, in contextual recommender systems, advertisements, news, movies or products are recommended to users based on the contexts: time, region, incoming users' age, gender, search query, purchase history, etc. Rewards of recommendation may be the click through rate (CTR), number of purchase, or a score derived from users' feedbacks (reviews, likes, dislikes, etc.). However, calculating the rewards may be costly, e.g., the costs of aggregating and analyzing users' feedbacks. Another example is the computer-aided medical diagnosis, where contexts are patients' profiles, disease history, symptoms, prior test results, etc. The diagnosis decision is whether or not to take some certain tests. However, taking certain tests is costly to the patients. Hence, judiciously making decisions about in what situations a test is taken is important.

Our proposed contextual MAB active learning (CMAB-AL) algorithm carries out two operations: the refinement of the context and arm spaces, and the selection of actions (both selection of arms and the action of requesting a ground truth). We divide the time into epochs, each of which contains a number of time slots. We assume that the context and arm spaces are metric spaces, and assume Lipschitz conditions on the expected rewards with respect to the metrics. In each epoch, we maintain a partition of the context space and the arm space. The partition of the spaces becomes finer and finer as the epoch grows. The selection of arms is based on the comparison of average reward difference and carefully selected thresholds. We use a strategic way to request the ground truth, ensuring that the estimations of rewards are accurate enough and hence the arms selected are near optimal. The goal is to minimize the regret, which is defined as the reward difference between the oracle strategy, where all information about the reward distribution is known, and the learning algorithm, where no knowledge about the distribution of rewards is known before learning. Previous works [10, 11, 12, 14, 15] show that the regret of MAB algorithms based on a metric space (context or arm space) can achieve the regret $R(T) = O(T^{\frac{d_X+d_A+1}{d_X+d_A+2}})$ ($d_X$ and $d_A$ are the covering dimensions[1] of the context space and the arm space). We analytically show that the regret of our proposed CMAB-AL algorithm is $R(T) = O(T^{\frac{d_X+d_A+1}{d_X+d_A+2}})$, which is sublinear in $T$ and in the same order with that of the conventional contextual multi-armed bandit algorithms, where no query cost is assumed.

Our main contributions are summarized as follows:

- We consider the contextual MAB framework with the stream-based online active learning setting, where requesting the ground truth of rewards is costly. Based on this framework, we propose the CMAB-AL algorithm to strategically making decisions about arm selections and query requests.
- We consider the scenario that not only the instance, but also prior information about the ground truth is provided to the annotator when requesting a ground truth. This prior information helps to reduce the query cost and to increase the learning efficiency.
- Through strategically design of requesting the ground truth, we analytically show that the regret of the proposed CMAB-AL algorithm can achieve the same order as that of conventional contextual MAB algorithms, where no query cost is assumed.

## 2 Problem Setup

### 2.1 Stream-based Active Learning Model

We consider the stream-based active learning model, where time is divided into discrete slots. The arm space is a bounded space $\mathcal{K}$, with covering dimension $d_A$. The context space is a bounded space $\mathcal{X}$, with covering dimension $d_X$. Given a context $x_t \in \mathcal{X}$ at time slot $t$, the reward of selecting arm $k_t \in \mathcal{K}$ is denoted by $r(x_t, k_t) \in [0, 1]$, which is sampled from some fixed but unknown distribution. The expected value of $r(x_t, k_t)$ is denoted by $\mu(x_t, k_t)$. For simplicity, we assume that the reward is in the interval $[0, 1]$. However, this can be easily extended to any bounded interval.

---

[1] The covering dimension of a metric space $\mathcal{X}$ is defined as the minimum of $d$, such that $\mathcal{N}(\mathcal{X}, d) \leq C_X \rho^{-d}$ holds for each $\rho > 0$, where $\mathcal{N}(\mathcal{X}, d)$ is the minimum number of balls with radius $\rho$ to cover the space $\mathcal{X}$, and $C_X$ is the covering constant for the space $\mathcal{X}$ [16].

We denote the action of requesting the ground truth of the reward by $q_t \in \{0, 1\}$, where $q_t = 1$ stands for requesting the ground truth, and $q_t = 0$ otherwise. We define the prior information about the reward $r(x_t, k_t)$ as a tuple $(a_t, b_t, \delta_t)$, which represents that the expected reward $\mu(x_t, k_t)$ is in the region $[a_t, b_t]$ with probability at least $1 - \delta_t$. The query cost is defined as a convex increasing function of the confidence interval $b_t - a_t$ and the significance level $\delta_t$, i.e., $c_t = c[(b_t - a_t)^{\beta_1} + \eta \delta_t^{\beta_2}]$, where $c > 0$ is a constant; $\beta_1, \beta_2 \geq 1$ are constants; and $\eta > 0$ is a trade-off factor to balance the two terms $(b_t - a_t)^{\beta_1}$ and $\delta_t^{\beta_2}$. The first term $(b_t - a_t)^{\beta_1}$ represents that a larger confidence interval $b_t - a_t$ results in more query cost, and the second term $\delta_t^{\beta_2}$ represents that a smaller confidence level $1 - \delta_t$ results in more query cost.

The process of online active learning at time slot $t$ is described as follows:

1. A context $x_t \in \mathcal{X}$ arrives.
2. An arm $k_t \in \mathcal{K}$ is selected by the algorithm.
3. A reward $r(x_t, k_t)$ is generated according to some fixed but unknown distribution. The algorithm chooses whether or not to request the ground truth of the reward. If $q_t = 1$, then the algorithm sends the context $x_t$, arm $k_t$, and prior information $(a_t, b_t, \delta_t)$ about the reward to the annotator, and an active learning cost $c_t$ is incurred. If $q_t = 0$, then there is no cost incurred and the reward cannot be observed. Hence, we denote the observation at time slot $t$ by $\hat{r}_t = r(x_t, k_t)$, if $q_t = 1$ and $\hat{r}_t = \emptyset$, if $q_t = 0$.

The history at time slot $t$ is defined as $h^{t-1} = \{(x_1, k_1, \hat{r}_1), (x_2, k_2, \hat{r}_2), \cdots, (x_{t-1}, k_{t-1}, \hat{r}_{t-1})\}$ for $t > 1$, and $h^0 = \emptyset$ for $t = 1$. The possible set of histories is defined as $\mathcal{H}$. We denote the set of algorithms by $\Pi$, and each algorithm $\pi \in \Pi$ is a mapping from the history and the current context to the arm selected and the action of whether to request the ground truth, namely, $\pi : \mathcal{H} \times \mathcal{X} \to \mathcal{K} \times \{0, 1\}$. To distinguish the arm selection and the action of whether to request the ground truth of the reward, we denote by $\pi_K^t = \pi_K(h^{t-1}, x_t)$ the arm selection at time slot $t$, and by $\pi_q^t = \pi_q(h^{t-1}, x_t)$ the action of whether to request the ground truth of the reward at time slot $t$.

## 2.2 Performance Evaluation

We use the total expected payoff (reward minus the query cost) up to time $T$ to describe the performance of the algorithm, denoted by $U_\pi(T) = E \sum_{t=1}^{T} [\mu(x_t, k_t) - c_t q_t]$. Clearly, there is an oracle algorithm $\pi^* \in \Pi$, which achieves the maximum payoff, namely, $\pi^* = \arg\max_{\pi \in \Pi} U_\pi(T)$ (Obviously, $\pi_q^*(h^{t-1}, x_t) = 0$ for all $t$). However, in practice, an algorithm needs to learn the best action corresponding to a context. Alternatively, we can use regret to describe the performance of an algorithm $\pi$, which is defined as:

$$R_\pi(T) = U_{\pi^*}(T) - U_\pi(T) \tag{1}$$

The goal of designing an algorithm $\pi$ is to minimize the regret $R_\pi(T)$.

As in [10, 11, 12], we assume that the rewards satisfy a Lipschitz condition with respect to both the context and the arm. Formally, we have the following assumption.

**Lipschitz Assumption**: For any two contexts $x, x' \in \mathcal{X}$ and two arms $k, k' \in \mathcal{K}$, we have the following Lipschitz conditions:

$$|\mu(x, k) - \mu(x', k)| \leq L_X ||x - x'||, \tag{2}$$

$$|\mu(x, k) - \mu(x, k')| \leq L_A ||k - k'||, \tag{3}$$

where $L_X, L_A$ are the Lipschitz constants for the context space and the arm space; $||x - x'||$ denotes the distance between two contexts $x$ and $x'$, and $||k - k'||$ denotes the distance between two arms $k$ and $k'$.

# 3 Active Learning

## 3.1 Active Learning Algorithm

In this subsection, we describe the active learning algorithm in the contextual bandit framework. Before describing the algorithm, we introduce some important notions.

- **Epoch, round, and slot**. A slot is the small time interval, corresponding to each $t$. A round $s$ is a varying time interval, and in each round, each active arm cluster has been selected just once. An epoch is the large time interval, which contains $T_i = 2^i$ time slots for epoch $i$. At the beginning of each epoch, the partitions of the arm space and the context space are updated.

- **Partition of the arm space and the context space**. Since the context space and the arm space are bounded continuous spaces, we partition the spaces into small subspaces at the beginning of each epoch, and each subspace is called an arm cluster or a context cluster. Formally, the partition of the context space is denoted by $\mathcal{P}_X(i) = \{\mathcal{X}_1, \mathcal{X}_2, \cdots, \mathcal{X}_{M_i}\}$, with cardinality $M_i = |\mathcal{P}_X(i)|$. The partition of the arm space is denoted by $\mathcal{P}_K(i) = \{\mathcal{K}_1, \mathcal{K}_2, \cdots, \mathcal{K}_{N_i}\}$, with cardinality $N_i = |\mathcal{P}_K(i)|$. The cluster radiuses [2] of the arm clusters and the context clusters are denoted by $\rho_{A,i}$ and $\rho_{X,i}$ in epoch $i$. We set the radiuses to be $\rho_{A,i} = \rho_{X,i} = T_i^{-\alpha}$, where $0 < \alpha < 1$.

- **Active arm cluster**. Only active arm clusters will be selected by the algorithm. At the beginning of each epoch, all arm clusters are set to be active. We denote the set of active arm clusters by $\mathcal{A}_m(i)$ for context cluster $\mathcal{X}_m$ in epoch $i$. As more knowledge is obtained through learning, some sub-optimal arm clusters are deactivated and will not be selected by the algorithm in the current epoch.

- **Control functions**. We set two control functions: $D_1(i, s_m(i))$ is used to deactivate sub-optimal arm clusters and $D_2(i, s_m(i))$ is used to check when to stop requesting the ground truth, where $s_m(i)$ is the number of rounds played so far in context cluster $\mathcal{X}_m$ in epoch $i$. When the stopping rule is satisfied, the algorithm stops deactivating arm clusters and runs in the exploitation phase in current epoch.

The CMAB-AL algorithm is described in Algorithm 1. The algorithm operates in epochs, and each epoch corresponds to a partition of the context and the arm space. The radiuses of the context and arm clusters become smaller as the epoch grows, implying a finer partition of the spaces. All arm clusters are set active at the beginning of each epoch. Each time a context arrives at a context cluster $\mathcal{X}_m$, an active arm cluster is selected, if it has not been selected in current round. If all active arm clusters are selected once, the current round ends and a new round begins. As more information about the reward accumulates, we deactivate the sub-optimal arm clusters based on a deactivating rule. Formally, we denote by $\bar{r}_{m,n}(s_m(i))$ the sample average reward in the context cluster $\mathcal{X}_m$ for arm cluster $\mathcal{K}_n$, and define the best arm cluster so far as $\bar{r}_m^*(s_m(i)) = \max_{n:\mathcal{K}_n \in \mathcal{A}_m(i)} \bar{r}_{m,n}(s_m(i))$. We calculate the sample average reward difference between best arm cluster so far and each active arm cluster $\bar{r}_m^*(s_m(i)) - \bar{r}_{m,n}(s_m(i))$, and compare this reward difference with a threshold $D_1(i, s_m(i))$, where $D_1(i, s_m(i)) = \varepsilon(i) + [4D(s_m(i)) + 2L_X \rho_{X,i} + 2L_A \rho_{A,i}]$; $\varepsilon(i) = LT_i^{-\alpha}$ ($L > 4L_X + 4L_A$ is a constant) is a small positive value for epoch $i$; and $D(s_m(i)) = \sqrt{\ln(2T_i^{1+\gamma})/2s_m(i)}$ ($0 < \gamma < 1$). If the sample average reward difference is greater than or equal to this threshold, the corresponding arm cluster has a high probability to be sub-optimal and is deactivated. When the reserved active arm clusters have sufficiently similar sample average rewards (i.e., they have a high probability to be near optimal), we stop the deactivating process based on a stopping rule. The stopping rule in epoch $i$ is based on the comparison of $\bar{r}_m^*(s_m(i)) - \bar{r}_{m,n}(s_m(i))$ with a threshold $D_2(i, s_m(i)) = 2\varepsilon(i) - [4D(s_m(i)) + 2L_X \rho_{X,i} + 2L_A \rho_{A,i}]$. When the sample average reward difference for any active arm cluster is smaller than or equal to this threshold, the deactivating process in current epoch stops. We denote by $S_m^i$ the number of rounds taken when the stopping rule is satisfied. In each round, when an arm cluster is selected, the context, the arbitrarily selected arm from that arm cluster, and the prior information are provided to the annotator to request a ground truth of the reward. The prior information is denoted by $(a_t, b_t, \delta_t)$, where $a_t = \bar{r}_{m,n}(s_m(i) - 1) - 2L_X \rho_{X,i} - 2L_A \rho_{A,i} - 2D(s_m(i) - 1)$,

---

[2]The radius of a cluster is defined as half the maximum distance between any two points in the cluster.

**Algorithm 1** The Contextual MAB based Active Learning Algorithm (CMAB-AL)
1: **for** epoch $i = 0, 1, 2 \cdots$ **do**
2:   **Initialization**: the partition of the context space $\mathcal{P}_X(i) = \{\mathcal{X}_1, \mathcal{X}_2, \cdots, \mathcal{X}_{M_i}\}$; the number of context clusters $M_i = |\mathcal{P}_X(i)|$; the partition of the arm space $\mathcal{P}_K(i) = \{\mathcal{K}_1, \mathcal{K}_2, \cdots, \mathcal{K}_{N_i}\}$; the number of arm clusters $N_i = |\mathcal{P}_K(i)|$; the set of active arm clusters with respect to context cluster $\mathcal{X}_m$: $\mathcal{A}_m(i) = \mathcal{P}_K(i)$, for all $1 \leq m \leq M_i$; the stop sign $stop_m = 0$, for all $1 \leq m \leq M_i$; the round counter $s_m(i) = 1$, for all $1 \leq m \leq M_i$; the average reward of selecting arm cluster $\mathcal{K}_n$: $\bar{r}_{m,n}(0) = 0$ for all $1 \leq m \leq M_i$ and $1 \leq n \leq N_i$.
3:   **for** time slot $t = 2^i$ to $2^{i+1} - 1$ **do**
4:     observe the context $x_t$ and find the context cluster $\mathcal{X}_m \in \mathcal{P}_m(i)$, such that $x_t \in \mathcal{X}_m$.
5:     **if** $stop_m = 0$, **then** {**Case 1: exploration**}
6:       Select an arm cluster $\mathcal{K}_n \in \mathcal{A}_m(i)$ that has not been selected in round $s_m(i)$, and arbitrarily select $k_t \in \mathcal{K}_n$.
7:       Choose $q_t = 1$, send the prior information $(a_t, b_t, \delta_t)$ to the annotator. (A query cost $c_t$ is incurred, and the reward $\hat{r}_t = r(x_t, k_t)$ is observed.)
8:       Update the average reward $\bar{r}_{m,n}(s_m(i)) = \frac{\bar{r}_{m,n}(s_m(i)-1)*(s_m(i)-1)+r(x_t,k_t)}{s_m(i)}$.
9:       **if** all arm clusters $\mathcal{K}_n \in \mathcal{A}_m(i)$ have been selected once in round $s_m(i)$, **then**
10:         **Deactivation**: check each arm cluster $\mathcal{K}_n \in \mathcal{A}_m(i)$, if $\bar{r}_m^*(s_m(i)) - \bar{r}_{m,n}(s_m(i)) \geq D_1(i, s_m(i))$, then deactivate the arm cluster $\mathcal{K}_n$, i.e., $\mathcal{A}_m(i) = \mathcal{A}_m(i) \setminus \{\mathcal{K}_n\}$.
11:         **Stop check**: if $\bar{r}_m^*(s_m(i)) - \bar{r}_{m,n}(s_m(i)) \leq D_2(i, s_m(i))$ for any $\mathcal{K}_n \in \mathcal{A}_m(i)$, then set the stop sign $stop_m = 1$.
12:         Update the round $s_m(i) = s_m(i) + 1$.
13:       **end if**
14:     **else** {$stop_m = 1$, **Case 2: exploitation**}
15:       Arbitrarily select $\mathcal{K}_n \in \mathcal{A}_m(i)$ and an arm in that cluster $k_t \in \mathcal{K}_n$.
16:       Choose $q_t = 0$. (The reward $r(x_t, k_t)$ is generated, but cannot be observed.)
17:     **end if**
18:   **end for**
19: **end for**

$b_t = \bar{r}_{m,n}(s_m(i) - 1) + 2L_X \rho_{X,i} + 2L_A \rho_{A,i} + 2D(s_m(i) - 1)$, $\delta_t = T_i^{-(1+\gamma)}$ for $s_m(i) > 1$, and $a_t = 0$, $b_t = 1$, $\delta_t = 0$ for $s_m(i) = 1$.

### 3.2 Regret Analysis

In this subsection, we show the performance of the proposed CMAB-AL algorithm, in terms of the regret. Before formally characterize the regret, we introduce several important notions.

- **Cluster reward**: We define the reward of selecting an arm cluster $\mathcal{K}_n$ in the context cluster $\mathcal{X}_m$ as $\mu(m, n) = \max_{x \in \mathcal{X}_m, k \in \mathcal{K}_n} \mu(x, k)$, and we define the optimal arm cluster with respect to the context cluster $\mathcal{X}_m$ as $\mu_m^* = \max_{1 \leq n \leq N_i} \mu(m, n)$. We define $\Delta_{m,n} = \mu_m^* - \mu(m, n)$.

- **$\varepsilon$-optimal arm cluster**: We define the $\varepsilon$-optimal arm cluster with respect to the context cluster $\mathcal{X}_m$ as the arm cluster $\mathcal{K}_n$ that satisfies $\mu(m, n) \geq \mu_m^* - \varepsilon$, and define the corresponding $\varepsilon$-suboptimal arm cluster as $\mathcal{K}_n$ that satisfies $\mu(m, n) < \mu_m^* - \varepsilon$.

- **Normal event and abnormal event**: we define the normal event for an arm cluster $\mathcal{K}_n$ in the context cluster $\mathcal{X}_m$ in round $s_m(i)$ in epoch $i$ as $\mathcal{N}_{m,n}(s_m(i)) = \{|\bar{r}_{m,n}(s_m(i)) - E[\bar{r}_{m,n}(s_m(i))]| \leq D(s_m(i))\}$, and the abnormal event as the complementary event set of $\mathcal{N}_{m,n}(s_m(i))$, i.e., $[\mathcal{N}_{m,n}(s_m(i))]^C$. If for an arm cluster $\mathcal{K}_n$ in the context cluster $\mathcal{X}_m$ in epoch $i$, no abnormal event occurs, then we denote this event by $\mathcal{N}_{i,m,n}$.

We introduce the following lemmas to characterize the properties of the algorithm.

**Lemma 1**: The abnormal event for an arm cluster $\mathcal{K}_n$ in epoch $i$ occurs with probability at most $\delta(i) = T_i^{-\gamma}$.

*Proof.* According to the definition of abnormal event and the Chernoff-Hoeffding bound, the probability that an abnormal event for an arm cluster occurs in round $s_m(i)$ can be bounded by

$$\Pr\{[\mathcal{N}_{m,n}(s_m(i))]^C\} \leq 2e^{-2[D(s_m(i))]^2 s_m(i)} \leq \frac{1}{T_i^{1+\gamma}}. \tag{4}$$

Hence, the probability that an abnormal event for an arm cluster $\mathcal{K}_n$ in epoch $i$ occurs with at most

$$\Pr\{[\mathcal{N}_{i,m,n}]^C\} \leq \sum_{s_m(i)=1}^{S_m^i} \Pr\{[\mathcal{N}_{m,n}(s_m(i))]^C\} \leq \sum_{s_m(i)=1}^{S_m^i} \frac{1}{T_i^{1+\gamma}} \leq \frac{1}{T_i^{\gamma}}. \tag{5}$$

□

**Lemma 2**: (a) With probability at least $1 - N_i \delta(i)$, $\varepsilon(i)$-optimal arm clusters are not deactivated in context cluster $\mathcal{X}_m$ in epoch $i$. (b) With probability at least $1 - N_i \delta(i)$, the active set $\mathcal{A}_m(i)$ in exploitation phases contains only $2\varepsilon(i)$-optimal arm clusters in context cluster $\mathcal{X}_m$ in epoch $i$.

*Proof.* If the normal event occurs, for any deactivated arm clusters $\mathcal{K}_n$ and $\mathcal{K}_{n'}$, we have:

$$\begin{aligned}
|\bar{r}_m^*(s_m(i)) - \bar{r}_{m,n}(s_m(i))| &\leq |E[\bar{r}_m^*(s_m(i))] - E[\bar{r}_{m,n}(s_m(i))]| \\
&+ |\bar{r}_m^*(s_m(i)) - E[\bar{r}_m^*(s_m(i))]| + |\bar{r}_{m,n}(s_m(i)) - E[\bar{r}_{m,n}(s_m(i))]| \\
&\leq |\mu_m^* - \mu(m,n)| + 2D(s_m(i)) + 2L_X \rho_{X,i} + 2L_A \rho_{A,i} + D(s_m(i)) + D(s_m(i))
\end{aligned} \tag{6}$$

Combining with the deactivating rule, we have $|\mu_m^* - \mu(m,n)| > \varepsilon(i)$. For any reserved arm clusters $\mathcal{K}_n$ and $\mathcal{K}_{n'}$, we have:

$$\begin{aligned}
|\bar{r}_{m,n}(S_m^i) - \bar{r}_{m,n'}(S_m^i)| &\geq |E[\bar{r}_{m,n}(S_m^i)] \\
&- E[\bar{r}_{m,n'}(S_m^i)]| - |\bar{r}_{m,n}(S_m^i) - E[\bar{r}_{m,n}(S_m^i)]| - |\bar{r}_{m,n'}(S_m^i) - E[\bar{r}_{m,n'}(S_m^i)]| \\
&\geq |\mu(m,n) - \mu(m,n')| - 2D(S_m^i) - 2L_X \rho_{X,i} - 2L_A \rho_{A,i} - D(S_m^i) - D(S_m^i)
\end{aligned} \tag{7}$$

Combining with the stopping rule, we have $|\mu(m,n) - \mu(m,n')| \leq 2\varepsilon(i)$. Since the normal event occurs with probability at least $1 - N_i \delta(i)$, the results follow. □

To bound the regret, we first consider the regret caused in context cluster $\mathcal{X}_m$ in epoch $i$, denoted by $R_{i,m}$. This regret can be decomposed into four terms: the regret $R_{i,m}^a$ caused by abnormal events, the regret $R_{i,m}^n$ caused by $2\varepsilon(i)$-optimal arm cluster selection and the inaccuracy of clusters, the regret $R_{i,m}^s$ caused by $2\varepsilon(i)$-suboptimal arm cluster selection when no abnormal events occur, and the query cost $R_{i,m}^q$. We have

$$R_{i,m} \leq R_{i,m}^a + R_{i,m}^n + R_{i,m}^s + R_{i,m}^q. \tag{8}$$

Let us denote by $T_i$ the number of time slots in epoch $i$, denote by $T_{i,m}$ the number of context arrivals in context cluster $\mathcal{X}_m$ in epoch $i$, and denote by $T_{i,m,n}$ the number of query requests for arm cluster $\mathcal{K}_n$ in context cluster $\mathcal{X}_m$ in epoch $i$. We set $\alpha = \frac{1}{d_A + d_X + 2}$ and $\gamma = \frac{d_A + 1}{d_A + d_X + 2}$.

For the first term $R_{i,m}^a$ in (8), when an abnormal event happens, the regret is $T_{i,m}$. According to Lemma 1, abnormal events happens with probability at most $\delta(i)$ for arm cluster $\mathcal{K}_n$ in epoch $i$. Therefore, the regret $R_{i,m}^a$ in (8) can be expressed as:

$$R_{i,m}^a \leq E \sum_{n=1}^{N_i} \sum_{t=2^i}^{2^{i+1}-1} T_{i,m} I\{\mathcal{N}_{m,n}(s_m(i))\} \leq \sum_{n=1}^{N_i} \delta(i) T_{i,m}. \tag{9}$$

For the second term $R_{i,m}^n$ in (8), the regret of $2\varepsilon(i)$-optimal arm cluster selection at each time slot is at most $2\varepsilon(i)$, and the regret of inaccuracy of clusters at each time slot is at most $2L_X \rho_{X,i} + 2L_A \rho_{A,i}$. Therefore, the regret $R_{i,m}^n$ can be expressed as:

$$R_{i,m}^n \leq \sum_{t=2^i}^{2^{i+1}-1} (2\varepsilon(i) + 2L_X \rho_{X,i} + 2L_A \rho_{A,i}) I\{x_t \in \mathcal{X}_m\} \leq 2(\varepsilon(i) + L_X \rho_{X,i} + L_A \rho_{A,i}) T_{i,m}. \tag{10}$$

For the third term $R_{i,m}^s$ in (8), when the normal event occurs, according to Lemma 2, $2\varepsilon(i)$-suboptimal arm cluster can only be selected in the exploration phases. Hence, the regret $R_{i,m}^s$ can be expressed as:

$$R_{i,m}^s \leq E \sum_{n:\Delta_{m,n}>2\varepsilon(i)} \sum_{t=2^i}^{2^{i+1}-1} \Delta_{m,n} I\{x_t \in \mathcal{X}_m, \pi_K^t \in \mathcal{K}_n, \pi_q^t = 1, \mathcal{N}_{i,m,n}\}. \quad (11)$$

According to the deactivating rule, the rounds of exploring arm cluster $\mathcal{K}_n$, $T_{i,m,n}$, can be bounded by the minimum $s$ that satisfies:

$$\Delta_{m,n} - 2D(s) - 2L_X \rho_{X,i} - 2L_A \rho_{A,i} \geq \bar{r}_m^*(s) - \bar{r}_{m,n}(s) \geq D_1(i,s). \quad (12)$$

Hence, for $\Delta_{m,n} > 2\varepsilon(i)$, we can bound $T_{i,m,n}$ by

$$T_{i,m,n} \leq \frac{18 \ln(2T_i^{1+\gamma})}{[\Delta_{m,n} - (\varepsilon(i) + 4L_X \rho_{X,i} + 4L_A \rho_{A,i})]^2}. \quad (13)$$

Therefore, the regret $R_{i,m}^s$ can be bounded by

$$\begin{aligned}
R_{i,m}^s &\leq E \sum_{n:\Delta_{m,n}>2\varepsilon(i)} \Delta_{m,n} T_{i,m,n} \\
&\leq E \sum_{n:\Delta_{m,n}>2\varepsilon(i)} \Big( \frac{18 \ln(2T_i^{1+\gamma})}{\Delta_{m,n}-(\varepsilon(i)+4L_X\rho_{X,i}+4L_A\rho_{A,i})} + \frac{18(\varepsilon(i)+4L_X\rho_{X,i}+4L_A\rho_{A,i})\ln(2T_i^{1+\gamma})}{[\Delta_{m,n}-(\varepsilon(i)+4L_X\rho_{X,i}+4L_A\rho_{A,i})]^2} \Big) \\
&\leq \frac{18N_i \ln(2T_i^{1+\gamma})}{2\varepsilon(i)-(\varepsilon(i)+4L_X\rho_{X,i}+4L_A\rho_{A,i})} + \frac{18N_i(\varepsilon(i)+4L_X\rho_{X,i}+4L_A\rho_{A,i})\ln(2T_i^{1+\gamma})}{[2\varepsilon(i)-(\varepsilon(i)+4L_X\rho_{X,i}+4L_A\rho_{A,i})]^2} \\
&\leq C_1 N_i \ln(2T_i^{1+\gamma}) T_i^\alpha
\end{aligned} \quad (14)$$

where $C_1 = \frac{36L}{(L-4L_X-4L_A)^2}$ is a constant.

For the fourth term $R_{i,m}^q$ in (8), we first consider the query cost $R_{i,m}^{q,1}$ when the abnormal event occurs. In this case, since the maximum query cost per slot is $2c$, the query cost can be bounded by

$$R_{i,m}^{q,1} \leq \sum_{n=1}^{N_i} 2c\delta(i) T_{i,m}. \quad (15)$$

Next, we consider the query cost $R_{i,m}^{q,2}$ in the case that only normal events occur. This can be bounded by

$$\begin{aligned}
R_{i,m}^{q,2} &\leq E \sum_{n=1}^{N_i} \sum_{t=2^i}^{2^{i+1}-1} c_t I\{x_t \in \mathcal{X}_m, \pi_K^t \in \mathcal{K}_n, \pi_q^t = 1\} \\
&\leq E \sum_{n=1}^{N_i} c + E \sum_{n=1}^{N_i} \sum_{s=2}^{S_m^i} [c(4L_X \rho_{X,i} + 4L_A \rho_{A,i} + 4D(s-1))^{\beta_1} + c\eta T_i^{-(1+\gamma)\beta_2}] \\
&\leq cN_i + cN_i \sum_{s=2}^{S_m^i} \frac{(8L_X\rho_{X,i}+8L_A\rho_{A,i})^{\beta_1}+(8D(s-1))^{\beta_1}}{2} + c\eta N_i \sum_{s=2}^{S_m^i} T_i^{-(1+\gamma)\beta_2}] \\
&\leq cN_i + cN_i 2^{3\beta_1-1}(L_X\rho_{X,i}+L_A\rho_{A,i})^{\beta_1} S_m^i + cN_i 2^{3\beta_1-1} \sum_{s=2}^{S_m^i} \frac{[\ln(2T_i^{\gamma+1})]^{\beta_1/2}}{2^{\beta_1/2}(s-1)^{\beta_1/2}} + c\eta N_i T_i^{-(1+\gamma)\beta_2+1}
\end{aligned} \quad (16)$$

where the third inequality is due to the Jensen's inequality. If $1 \leq \beta_1 < 2$, the third term on the right hand side of the last inequality in (16) can be bounded by $cN_i 2^{5\beta_1/2-1} \frac{[\ln(2T_i^{\gamma+1})]^{\beta_1/2}(S_m^i)^{1-\beta_1/2}}{1-\beta_1/2}$, due to the divergent series $\sum_{t=1}^T t^{-y} \leq T^{(1-y)}/(1-y)$ for $0 < y < 1$ [17]. If $\beta_1 \geq 2$, the third term on the right hand side of the last inequality in (16) can be bounded by $cN_i 2^{5\beta_1/2-1}[\ln(2T_i^{\gamma+1})]^{\beta_1/2}(\ln S_m^i)^{\beta_1/2}$, due to the series $\sum_{t=1}^{T-1} t^{-y} \leq \ln T$ for $y \geq 1$. We can also have the bound of $S_m^i$ due to the fact that when $D_1(i, S_m^i) \leq D_2(i, S_m^i)$, the stopping rule is satisfied. Hence, $S_m^i$ can be bounded by the minimum $s$, such that $D_1(i,s) \leq D_2(i,s)$. This shows:

$$S_m^i \leq \frac{64 T_i^{2\alpha} \ln(2T_i^{\gamma+1})}{(L-4L_X-4L_A)^2}. \quad (17)$$

Thus, we can bound $R_{i,m}^{q,2}$ by

$$R_{i,m}^{q,2} \leq \begin{cases} C_2 N_i T_i^{\alpha(2-\beta_1)} \ln(2T_i^{\gamma+1}), & if\ 1 \leq \beta_1 < 2 \\ C_3 N_i [\ln(2T_i^{\gamma+1})]^{\beta_1}, & if\ \beta_1 \geq 2 \end{cases}, \quad (18)$$

where $C_2 = c(1+\eta) + \frac{c2^{3\beta_1+5}(L_X+L_A)^{\beta_1}}{(L-4L_X-4L_A)^2} + \frac{c2^{6-\beta_1/2}}{(2-\beta_1)(L-4L_X-4L_A)^2}$ is a constant, $C_3 = c(1+\eta+2^{5\beta_1/2-1}) + \frac{c2^{3\beta_1+5}(L_X+L_A)^{\beta_1}}{(L-4L_X-4L_A)^2}$ is a constant.

Then we consider the regret bound of the CMAB-AL algorithm by adding all the context clusters and all the epochs, and get the following theorem.

**Theorem 1**: The regret of the CMAB-AL algorithm can be bounded by $R(T) = O(T^{\frac{d_X+d_A+1}{d_X+d_A+2}})$.

*Proof.* The regret of the CMAB-AL algorithm can be bounded by

$$R(T) \leq E \sum_{i=0}^{\log_2 T} \sum_{m=1}^{M_i} R_m^i. \quad (19)$$

According to the definition of covering dimensions [16], the maximum number of arm clusters can be bounded by $N_i \leq C_A \rho_{A,i}^{-d_A}$ in epoch $i$, and the maximum number of context clusters can be bounded by $M_i \leq C_X \rho_{X,i}^{-d_X}$ in epoch $i$, where $C_A, C_X$ are covering constants for the arm space and the context space. We also note that $\sum_{m=1}^{M_i} T_{i,m} = T_i$. Hence, the regret can be bounded by

$$\begin{aligned} R(T) &\leq E \sum_{i=0}^{\log_2 T} \sum_{m=1}^{M_i} R_m^i \\ &\leq E \sum_{i=0}^{\log_2 T} C_X \rho_{X,i}^{-d_X} C_A \rho_{A,i}^{-d_A} [(\delta(i) + 2c\delta(i) + 2\varepsilon(i) + 2L_X \rho_{X,i} + 2L_A \rho_{A,i}) T_i \\ &\qquad\qquad\qquad\qquad + C_1 \ln(2T_i^{1+\gamma}) T_i^\alpha + C_2 T_i^\alpha \ln(2T_i^{\gamma+1})] \\ &\leq \sum_{i=0}^{\log_2 T} C' T_i^{\alpha(d_X+d_A+1)} \ln(T_i) \\ &\leq C T^{\frac{d_X+d_A+1}{d_X+d_A+2}} \ln(T) \end{aligned} \quad (20)$$

where $C'$ and $C$ are constants. Therefore, the result of Theorem 1 follows. □

We can see from Theorem 1 that this regret matches the order of the conventional contextual MAB algorithm [10,12,14], where no query cost is assumed when the ground truth is requested.

### 3.3 A Lower Bound

In this subsection, we show a lower bound for the CMAB-AL algorithm. Since the proposed algorithm incurs the query cost when it requests a ground truth, the lower bound of the regret cannot be lower than that of the conventional contextual MAB setting where no query cost is incurred.

***Remark***: A lower bound of the CMAB-AL algorithm is $\Omega(T^{\frac{d_X+d_A+1}{d_X+d_A+2}})$, which can be directly derived from the lower bound results of conventional contextual MAB [10,12].

## 4 Conclusions

In this paper, we propose a stream-based online active learning algorithm in a contextual MAB framework. We consider the varying active learning cost changing with prior information provided to the annotator. The algorithm maintains a partition of the context space and the arm space for a certain time, and chooses the arm and whether or not to request a ground truth of the reward. Through precise control of the partitions of the spaces and the selection of arms and when to request a ground truth of the reward, the algorithm can balance the accuracy of learning and the cost incurred by active learning. We analytically show that the regret of the proposed algorithm can achieve the same order as that of conventional contextual MAB algorithms, where no query cost is assumed.